\documentclass[10pt,twocolumn,letterpaper]{article}

\usepackage{cvpr}              % To produce the CAMERA-READY version
\usepackage{graphicx}
\usepackage{amsmath}
\usepackage{amssymb}
\usepackage{booktabs}
\usepackage{bbding}

\usepackage{stfloats}
\usepackage{multirow}

\usepackage[pagebackref,breaklinks,colorlinks]{hyperref}

\usepackage[capitalize]{cleveref}
\crefname{section}{Sec.}{Secs.}
\Crefname{section}{Section}{Sections}
\Crefname{table}{Table}{Tables}
\crefname{table}{Tab.}{Tabs.}

\def\cvprPaperID{*****} % *** Enter the CVPR Paper ID here
\def\confName{CVPR}
\def\confYear{2022}

\begin{document}

\title{STU-Net: Scalable and Transferable Medical Image Segmentation Models Empowered by Large-Scale Supervised Pre-training}

\author{
Ziyan Huang\textsuperscript{1,2}\thanks{Equal contribution. This work is done when Ziyan Huang is an intern at Shanghai AI Laboratory.}
\qquad Haoyu Wang\textsuperscript{1,2}\footnotemark[1]
\qquad Zhongying Deng\textsuperscript{2}\footnotemark[1]
\qquad Jin Ye\textsuperscript{2}\footnotemark[1]
\qquad Yanzhou Su\textsuperscript{2}\\
Hui Sun\textsuperscript{2}
\qquad Junjun He\textsuperscript{2}
\qquad Yun Gu\textsuperscript{1} 
\qquad Lixu Gu\textsuperscript{1} 
\qquad Shaoting Zhang\textsuperscript{2} 
\qquad Yu Qiao\textsuperscript{2}\thanks{Corresponding author}\\
\\
\textsuperscript{1}Shanghai Jiao Tong University\qquad \\ \textsuperscript{2}Shanghai AI Laboratory\\
{\tt\small \{ziyanhuang, gulixu\}@sjtu.edu.cn}\\
{\tt\small \{hejunjun, yejin, zhangshaoting, qiaoyu\}@pjlab.org.cn}
}
\maketitle

%%%%%%%%% ABSTRACT
\begin{abstract}
Large-scale models pre-trained on large-scale datasets have profoundly advanced the development of deep learning. However, the state-of-the-art models for medical image segmentation are still small-scale, with their parameters only in the tens of millions. Further scaling them up to higher orders of magnitude is rarely explored. An overarching goal of exploring large-scale models is to train them on large-scale medical segmentation datasets for better transfer capacities. In this work, we design a series of Scalable and Transferable U-Net (STU-Net) models, with parameter sizes ranging from 14 million to 1.4 billion. Notably, the 1.4B STU-Net is the largest medical image segmentation model to date.  
Our STU-Net is based on nnU-Net framework due to its popularity and impressive performance. We first refine the default convolutional blocks in nnU-Net to make them scalable. Then, we empirically evaluate different scaling combinations of network depth and width, discovering that it is optimal to scale model depth and width together. We train our scalable STU-Net models on a large-scale TotalSegmentator dataset and find that increasing model size brings a stronger performance gain. This observation reveals that a large model is promising in medical image segmentation. 
Furthermore, we evaluate the transferability of our model on 14 downstream datasets for direct inference and 3 datasets for further fine-tuning, covering various modalities and segmentation targets. We observe good performance of our pre-trained model in both direct inference and fine-tuning. The code and pre-trained models are available at \href{https://github.com/Ziyan-Huang/STU-Net}{https://github.com/Ziyan-Huang/STU-Net}.
\end{abstract}

\begin{figure}[t]
  \centering
   \includegraphics[width=\linewidth]{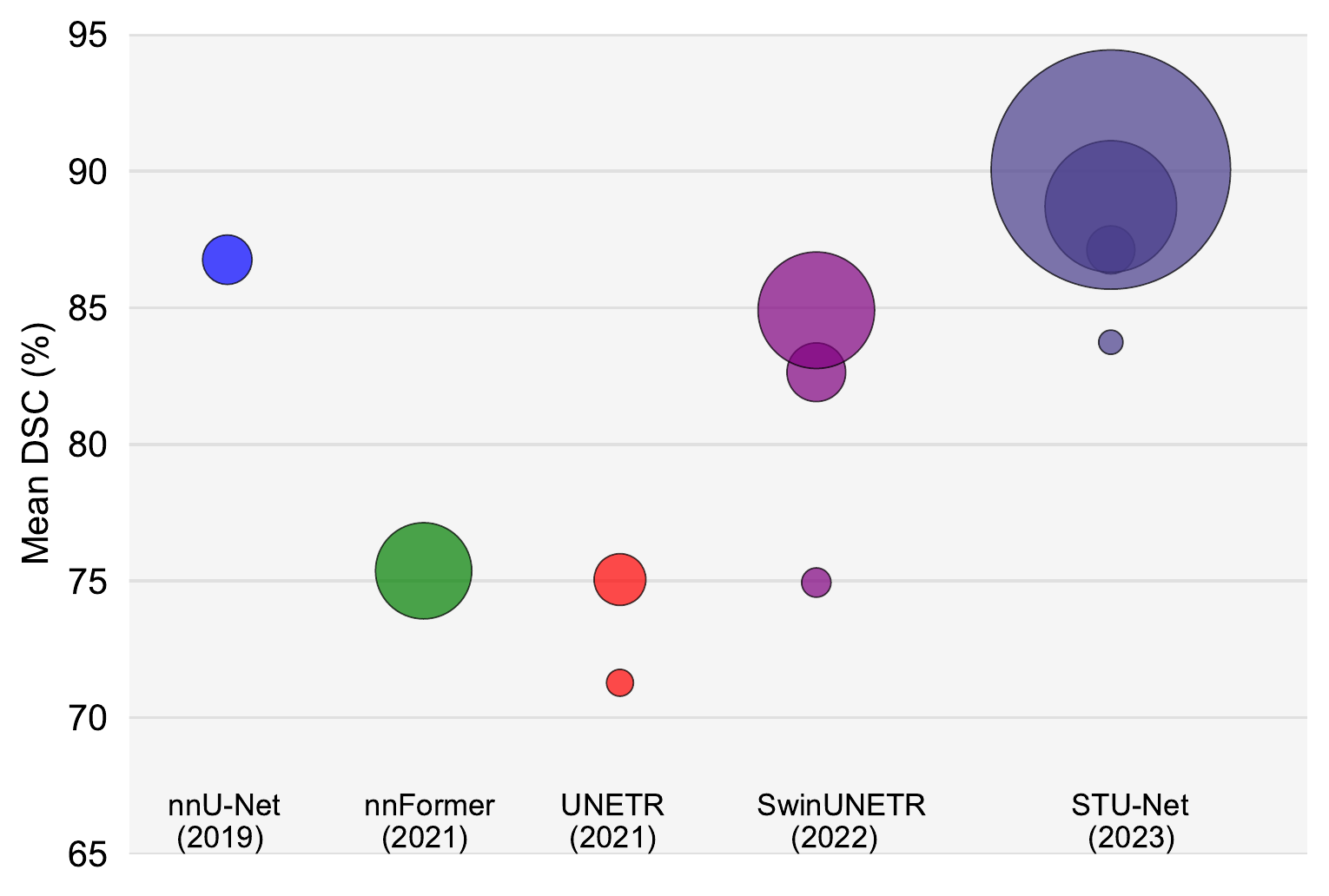}
   \caption{Segmentation performance of various models on the TotalSegmentator dataset. The area of each bubble is proportional to the FLOPs (Floating-Point Operations Per Second) of the corresponding model at different scales. Distinct colors represent different models, while multiple bubbles of the same color denote the same model with varying scales. FLOPs calculations are based on input patch sizes of $128\times 128\times 128$.}
   \label{fig:bubble}
\end{figure}

%%%%%%%%% BODY TEXT
\section{Introduction}
\label{sec:intro}

Medical image segmentation, which aims to automatically annotate anatomical structures and lesions in medical images, is an important intermediate step for many downstream clinical tasks, such as medical image registration \cite{wu2022car}, quantification \cite{li2022medical} and image-guided surgery \cite{wang2022integrated}.   
In recent years, various specific medical image segmentation tasks have been heavily studied and great success has been achieved by many deep learning based models \cite{ronneberger2015u, hatamizadeh2022unetr, hatamizadeh2022swin, zhou2021nnformer}. 
However, these models often require careful tuning to adapt to different tasks, limiting their transferability. There is a lack of work investigating a single model to simultaneously handle various medical segmentation tasks, including various input modalities (CT, MRI, PET, etc.) and various segmentation targets, such as organs and tumors. 

The key to such a goal can be pre-training large-scale models on large-scale datasets to make the models \textbf{transferable}. This solution has been  verified in both vision and language fields \cite{brown2020language, dehghani2023scaling,kaplan2020scaling,baltruschat2021scaling,kirillov2023segment}, and its success can be explained as follows: large-scale datasets can provide more knowledge for model training, while large-scale models with more parameters can better take advantage of this knowledge. From the perspective of datasets, some public large-scale medical image segmentation datasets are emerging. For example, AbdomenCT-1K \cite{ma2021abdomenct}, BraTS21 \cite{baid2021rsna}, AutoPET \cite{gatidis2022whole}, and TotalSegmentator \cite{wasserthal2022totalsegmentator} contain more than 1000 annotated image from diverse modalities (including CT, PET, MRI) and segmentation targets (abdomen organs, brain/whole-body tumor). Notably, the large-scale TotalSegmentator \cite{wasserthal2022totalsegmentator} dataset has 1204 CT images with 104 organs annotated, which is suitable for training a large-scale model. 

With the availability of large-scale datasets for pre-training, our primary focus is on designing large-scale models that are \textbf{transferable} to a variety of medical segmentation tasks. Nonetheless, large-scale models usually take much more computational costs. This can be even severer when 3-dimensional high-resolution medical images are used for training. Hence, we also hope that the large-scale model can be \textbf{scalable} to different sizes to fit different computational budgets.

To sum up, our goal is to design a large-scale medical segmentation model that is \textbf{scalable and transferable}. To this end, we propose a series of Scalable and Transferable U-Net \cite{ronneberger2015u}, termed STU-Net, with parameter sizes ranging from 14 million to 1.4 billion. It is worth noting that the 1.4B model is the largest model in the medical image segmentation field to date. In addition to these different model sizes for scalability, we pre-train them on large-scale datasets in a supervised manner to ensure the models' transfer capacities.
Specifically, we build our models based on nnU-Net framework~\cite{isensee2021nnu} due to its state-of-the-art baseline performance and its wide use by researchers. 
However, there are two obstacles to developing large-scale models using this framework. 1) For the models' scalability, the basic convolutional blocks in nnU-Net may not be suitable for scaling due to gradient diffusion, and how to increase parameters remains unclear. 2) To evaluate the models' transferability, we usually need to fine-tune the models on other downstream datasets. But the model architectures in nnU-Net cannot easily be used for fine-tuning since they are treated as hyper-parameters and thus task-specific.

To tackle the first obstacle, we refine the basic convolutional blocks of nnU-Net, such as incorporating residual connection \cite{he2016deep} to its basic blocks, to facilitate scaling model depth. Then we empirically evaluate different combinations of network depth and width. We discover that it is optimal to scale model depth and width together. Hence, we can obtain the STU-Net-L (2× depth, 2× width) with 450 million parameters and the STU-Net-H (3× depth, 3× width) with 1.4 billion parameters. We do not conduct further scaling due to the prohibitive computational costs and GPU memory consumption on a single A100 GPU.

For the second obstacle, we replace the transpose convolution in nnU-Net with the nearest interpolation followed by a $1\times 1\times 1$ convolution for up-sampling, which avoids the use of task-specific kernel/stride options in transpose convolution and makes up-sampling blocks transferable. We further fix hyper-parameters related to model weights, e.g., fixing the number of stages to 6 and using isotropic convolution kernels, so that the model architecture can be the same during pre-training and fine-tuning, thus feasible for transfer to other tasks.
 
To make our STU-Net transferable, we further adopt the large-scale TotalSegmentator dataset for supervised pre-training. TotalSegmentator is one of the largest datasets in terms of the number of training images and categories, containing 1204 volumetric CT images with 104 well-annotated structures in the whole body. The large-scale STU-Net pre-trained on the large-scale dataset excels in various downstream datasets, covering diverse modalities and segmentation targets, when doing direct inference and fine-tuning. 
The superior performance shows that our large-scale STU-Net has impressive transferability, benefiting from the large-scale supervised pre-training.

Our contribution can be summarized as:
\begin{itemize}
    \item 
    We propose scalable STU-Net models which can be scaled to different parameter sizes, with the 1.4B STU-Net as the largest medical image segmentation model to date. Based on these STU-Net models, we discover that model's performance significantly increases with respect to model size when trained on large-scale datasets.
    \item 
    Our large-scale STU-Net model pre-trained on large-scale TotalSegmentator dataset demonstrates strong transfer capabilities. It can generalize well to a variety of other datasets without additional tuning or adaptation. It also excels on downstream datasets with fine-tuning. 
    \item 
    Our STU-Net is based on the nnU-Net framework and benefits from its task-adaptive designs, so our model can adapt to different tasks with good performance guarantee. Meanwhile, we modify the model architecture in nnU-Net to ensure that model weights trained on the TotalSegmentator dataset can be easily transferred to downstream tasks.
\end{itemize}

%-------------------------------------------------------------------------
\section{Related Works}
\subsection{Medical Image Segmentation Models}
Medical image segmentation is dominated by deep learning models, which can be broadly divided into two categories: CNN-based models and Transformer-based models. U-Net \cite{ronneberger2015u} is the pioneering CNN model proposed for medical image segmentation. Based on it, residual connection \cite{milletari2016v, drozdzal2016importance}, attention module \cite{oktay2018attention} and different feature aggregation strategies \cite{zhou2019unet++} are applied for various tasks. Recently, vision transformers \cite{dosovitskiy2020vit, liu2021swin} with self-attention mechanisms \cite{vaswani2017attention}, which achieve success in natural image processing, are introduced in medical image segmentation tasks. Particularly, UNETR \cite{hatamizadeh2022unetr} and SwinUNETR \cite{hatamizadeh2022swin} use Vision Transformer and Swin Transformer respectively as encoders to extract features from embedded 3D patches and positional embedding. TransUNet \cite{chen2021transunet} uses transformer blocks as a bottleneck to extract global contexts. nnFormer \cite{zhou2021nnformer} proposes an interleaved combination of transformer and convolution blocks to extract both local and global features. These medical image segmentation models have only several million parameters, thus not large enough. In addition, these models are not scalable and transferable to fit different computational budgets and handle diverse medical image segmentation tasks simultaneously.

\subsection{Scaling-up Models}
Scaling up models is widely used to boost models' performance in deep learning. The most common way is to scale the depth \cite{he2016deep} and width \cite{zagoruyko2016wide} of models. EfficientNet \cite{tan2019efficientnet} proposes to scale network depth, width, and resolution in a compound way. \cite{kaplan2020scaling} and \cite{zhai2022scaling} present comprehensive studies of empirical scaling laws of transformers in language processing and vision recognition respectively, where the main finding shows that the relationships between computation, data size, model size, and performance fit a power law. Following the scaling laws, GPT-3 \cite{brown2020language}, which has 175 billion parameters and is pre-trained on about 45 TB text data, has near-human performance in various text processing tasks. In addition, vision Transformers \cite{dosovitskiy2020vit} have been scaled up to 22 billion parameters \cite{dehghani2023scaling} and pre-trained on approximately 4 billion images \cite{zhai2022scaling}\cite{sun2017revisiting}. With a frozen visual feature extractor, the ViT-22B model achieved a top-1 accuracy of 89.5\% on the ImageNet \cite{deng2009imagenet} dataset.
However, only a few works scale models in medical image segmentation to large scale. Following EfficientNet, \cite{baltruschat2021scaling} scales 2D U-Net for biodegradable bone implant segmentation. \cite{huang2022adwu} explores task-specific scaling strategies for various medical image segmentation tasks. These works have limited scalability and are only evaluated on small datasets. In contrast, our paper successfully scales models to be an order of magnitude larger than previous works and evaluates their transfer capacities on large-scale datasets.
%-------------------------------------------------------------------------

\section{Methods}
We build our models based on the nnU-Net framework which can configure task-specific hyper-parameters automatically and achieve state-of-the-art performance on various tasks. In this section, we first introduce our refinements on it to facilitate scalability and transferability, then present our proposed scaling method, and finally detail our large-scale supervised pre-training strategy for better transferability.

\subsection{nnU-Net Architecture}
We first briefly introduce the default nnU-Net architecture, especially its modules related to our modifications. nnU-Net adopts a symmetric encoder-decoder architecture based on skip connections. Such an architecture contains various resolution stages. Each stage comprises two convolution layers followed by instance normalization and leaky ReLU (denoted as Conv-IN-LeakyReLU). It does not contain residual connections, so simply stacking more layers in each stage may suffer from gradient diffusion, making the whole model hard to optimize. This can limit the depth of nnU-Net, further constraining the scalability of nnU-Net. 
On the other hand, nnU-Net determines the input patch size and input spacing according to dataset properties. Then, the dataset-sensitive patch size and spacing are used to set the hyper-parameters related to network architecture, like the number of resolution stages, convolution kernels and down-sample/up-sample ratios. Thus, these architecture-related hyper-parameters vary between tasks, leading to different network architectures for different tasks. As such, the models trained on one task cannot be transferred to the other tasks directly, which restricts the evaluation of models' transfer capacities.

\begin{table*}[htbp]
\caption {Comparison of different hyper-parameters in nnU-Net, 3D U-Net and our method. The up(down)-sample ratios along different axes are (x,y,z). nnU-Net configures task-specific hyper-parameters automatically, which usually brings state-of-the-art performance. But this may make the model less transferable. The 3D-UNet uses a fixed network architecture that can easily transfer weights across different segmentation tasks but its baseline performance is usually worse than nnU-Net. Our STU-Net not only maintains good performance of nnU-Net but also can be transferred easily.} \label{table:setting}
\centering
{\begin{tabular}{c|c|c|c}
\hline
Settings &  nnU-Net &  3D U-Net  &  STU-Net (ours) \\ \hline
number of resolution stages   &   4-7   &   5  &   6 \\
convolution kernels   &   3$\times$3$\times$3 or 3$\times$3$\times$1  &   3$\times$3$\times$3 & 3$\times$3$\times$3\\
up(down)-sample ratios & (2,2,2) or (2,2,1) & (2,2,2) & (2,2,2) or (2,2,1) \\
input patch size  & task-specific &   fixed   &   task-specific \\
input spacing & task-specific   &   fixed &   task-specific\\
up-sample operation & transpose convolution & transpose convolution & interpolation with (1,1,1) convolution \\
\hline
\end{tabular}}
\end{table*}

\subsection{nnU-Net Tweaks}
The task-specific hyper-parameters in nnU-Net can be divided into model weight-related (e.g., convolution kernel size, number of resolution stages) and  weight-unrelated (such as pool kernels, input patch size, and input spacing). We first fix weight-related hyper-parameters to keep the model architecture feasible for transfer to other tasks. Concretely, we keep the number of resolution stages in all tasks to be 6, and use isotropic kernels (3,3,3) for all convolution layers. For weight-unrelated hyper-parameters, we adopt the default setting in nnU-Net to ensure its state-of-the-art performance. We compare our settings with nnU-Net and 3D U-Net \cite{zhang2021dodnet} in Table \ref{table:setting}.

\begin{figure}[tb]
  \centering
   \includegraphics[width=\columnwidth]{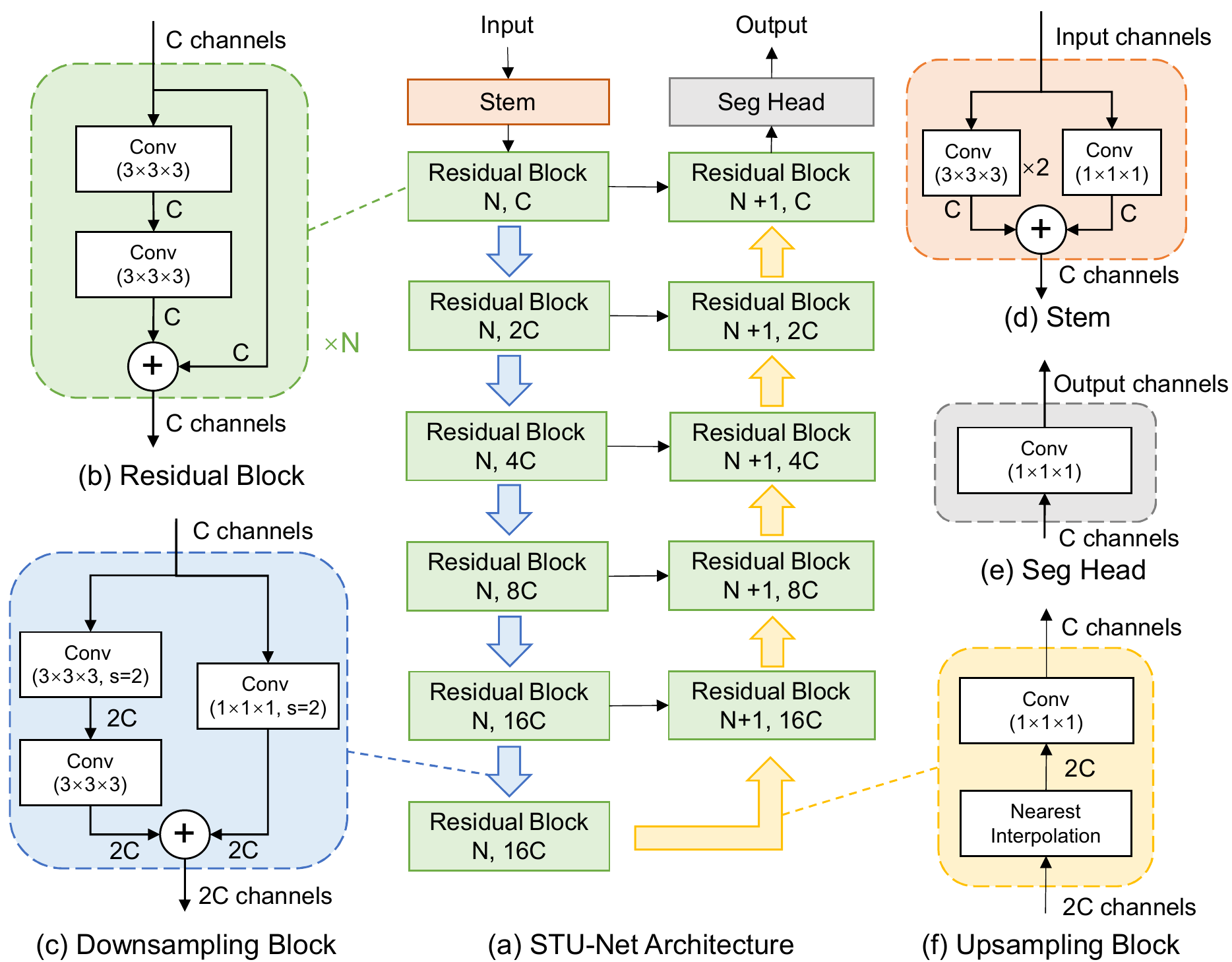}
   \caption{Illustration of our STU-Net architecture which is built upon the nnU-Net architecture with several modifications to enhance its scalability and transferability.
   (a) An overview of the STU-Net architecture. The blue arrows denote downsampling while the yellow ones represent upsampling. (b) Residual blocks to achieve a large-scale model. (c) Downsampling in the first residual block of each encoder stage. (d-e) Stem and segmentation head for channel conversion of input and output. (f) Weight-free interpolation for upsampling, which effectively addresses the issue of weight mismatch across different tasks. }
   \label{fig:model}
\end{figure}

\subsubsection{Basic Block Tweaks}
In both the encoder and decoder of nnU-Net, each stage consists of a basic block. Each basic block comprises two Conv-IN-LeakyReLU layers. When increasing the number of basic blocks in each stage, optimization issues may happen due to gradient diffusion. To address these issues, we introduce residual connections in the basic block. Apart from the first resolution stage, each stage in the encoder begins with a downsampling block in Figure~\ref{fig:model}(c), followed by several residual blocks in Figure~\ref{fig:model}(b). Different from nnU-Net which uses a separate convolution for downsampling, we integrate downsampling in the first residual block of each stage. Such a downsampling block has two branches, namely the left and right branches as in Figure~\ref{fig:model}(c). The left branch has two $3\times 3\times 3$ convolutions with different strides, namely a stride of 1 for the former one and 2 for the latter. The right branch uses a kernel size $1\times 1\times 1$ with a stride of 2 to match the output shape of the left branch. This downsampling block is with similar residual architecture to the regular residual block in Figure~\ref{fig:model}(b), making the whole architecture neat.

\subsubsection{Upsampling Tweaks}
By default, nnU-Net uses transpose convolution with stride for upsampling. However, the convolutional kernels and strides may vary between (2,2,2) and (2,2,1) for different tasks even in the same resolution stage, which causes different weight shapes in the transpose convolution for different tasks. It further leads to a weight mismatch when we want to transfer the weights from one task to another. To solve this problem, we use interpolation followed by a $1\times 1\times 1$ convolution with stride 1 to replace the transpose convolution, as the weight-free interpolation has no weight shape issue (see Figure~\ref{fig:model}(f)). We use nearest neighbor interpolation for up-sampling based on our experimental results (see Table \ref{table:ablation_arch}), which show that nearest neighbor interpolation not only offers faster processing but also achieves comparable performance to cubic linear interpolation.

\subsection{Scaling Strategy}
Based on the refined modules presented above, we can obtain a new network architecture called STU-Net. We further expand the depth and width in each stage of our model to scale it up. Deeper networks usually have larger receptive fields and better representation abilities. Wider networks tend to extract richer multi-scale features in each layer. 
As revealed in EfficientNet \cite{tan2019efficientnet}, depth scaling and width scaling are not independent. It is desirable to scale the depth and width of a network in a compound way to achieve better accuracy and efficiency. To simplify the scaling problem, we maintain the symmetry of our model, which means that we scale the encoder and decoder simultaneously and scale the depth and width with the same ratio in each resolution stage. Table~\ref{table:model} displays the different scales of STU-Net, with the suffixes 'S, B, L, H' representing 'Small, Base, Large, and Huge,' respectively.

\begin{table}[tb]
\caption{Our proposed STU-Net with different scales. Depth refers to the number of residual blocks in each resolution stage, and width denotes the channel count in each resolution stage. Parameter calculations are based on a single-channel input and a 105-channel output, which accounts for 104 foreground classes and 1 background class in the TotalSegmentator dataset. FLOPs calculations are based on input patch sizes of 128 x 128 x 128.}
\label{table:model}
\centering
\resizebox{\columnwidth}{!}{
{\begin{tabular}{c|cc|cc}
\hline
 Model &  depth &  width  &  Params (M) & FLOPs (T)  \\ \hline
STU-Net-S & (1,1,1,1,1,1) & (16,32,64,128,256,256) & 14.60 & 0.13 \\
STU-Net-B & (1,1,1,1,1,1) & (32,64,128,256,512,512) & 58.26 & 0.51 \\
STU-Net-L & (2,2,2,2,2,2) & (64,128,256,512,1024,1024) & 440.30 & 3.81 \\
STU-Net-H & (3,3,3,3,3,3) & (96,192,384,768,1536,1536) & 1457.33 & 12.60    \\
\hline
\end{tabular}}
}
\end{table}

\subsection{Large-Scale Supervised Pre-training}
We pre-train our STU-Net on the TotalSegmentator dataset. In STU-Net, the final $1\times 1\times 1$ convolution layer for segmentation output has 105 channels, which corresponds to the total number of target annotation categories in TotalSegmentator. 
To make the pre-trained models more general and transferable, we do not strictly follow the standard training procedure in nnU-Net but make some modifications. 
Compared to the default 1000 training epochs in nnU-Net, we pre-train our models for 4000 epochs. In addition, we discover that pre-training with mirror data augmentation can improve transfer performance on downstream tasks. 

Our pre-trained models can directly perform inference on downstream datasets that consist of CT images with target segmentation categories within the upstream 104 classes, without further tuning. If the downstream tasks have novel labels or different modalities, we use our trained model as initialization  and randomly initialize the segmentation output layer to match the number of target output classes. 
For fine-tuning, the segmentation head is randomly initialized while the weights of the remaining layers are loaded from our pre-trained model. These weights are fine-tuned with a smaller learning rate (0.1$\times$) than that of the segmentation head, which leads to better results.

\section{Experiments}
\noindent\textbf{Dataset}
We train our STU-Net of different scales on the TotalSegmentator \cite{wasserthal2022totalsegmentator} dataset which contains 1204 images with 104 anatomical structures (consisting of 27 organs, 59 bones, 10 muscles and 8 vessels). It covers most of the clinical segmentation targets for normal structures in the whole body. All the images are resampled to $1.5\times 1.5 \times 1.5 $ mm isotropic resolution. We follow the original data split in \cite{wasserthal2022totalsegmentator} that uses 1081 cases for training, 57 cases for validation, and 65 cases for final testing.  It is noticeable that faces have been blurred for data privacy reasons. We evaluate our trained STU-Net on 14 public datasets for direct inference and 3 public datasets for further fine-tuning to test the transferability of our trained models. The detailed properties of these downstream datasets are shown in Appendix.

\noindent\textbf{Evaluation Metric}
We adopt the Dice Similarity Coefficient (DSC) as the evaluation metric and a higher DSC score indicates better segmentation performance. For a fair comparison, we report the results of models training at the last epoch instead of the best one. 

\noindent\textbf{Implementation Details}
We run all the experiments based on the environment of Python 3.8, CentOS 7, Pytorch 1.10, and nnU-Net 1.7.0. 
We roughly follow the default data pre-processing, data augmentation and training procedure in nnU-Net. We use SGD optimizer with Nestrov momentum of 0.99, and a weight decay of 1e-3. The batch size is fixed to 2 and each epoch contains 250 iterations. For all datasets, the learning rate starts at 0.01 when training from scratch, except for AutoPET \cite{gatidis2022whole}, which begins at 0.001, following the state-of-the-art solution \cite{ye2022exploring}.  The learning rate is decayed following the poly learning rate policy: $(1-epoch/1000)^{0.9}$. We adopt data augmentation of additive brightness, gamma, rotation, scaling, mirror, and elastic deformation on the fly during training. The pre-training patch size on TotalSegmentator is $128\times 128\times 128$. Fine-tuning patch sizes on downstream tasks are configured by nnU-Net automatically. Models are trained on NVIDIA Tesla A100 cards with 80 GB VRAM.

\begin{table*}[htbp]
\caption {Segmentation results of different methods on TotalSegmentator validation dataset. Mean DSC ($\% \uparrow$) is evaluated. Due to the limited space, we divide all 104 classes into 5 groups and report the mean DSC of these 5 group classes and all classes, respectively. *: We change the maximum feature number in nnU-Net from 320 to 512 to match the parameters of our STU-Net-B model. $\dagger$: SwinUNETR-L is obtained by changing the feature size in SwinUNETR-B from 48 to 96.}\label{table:TotalSeg_Results}
\centering
\resizebox{\linewidth}{!}{
{\begin{tabular}{c|cc|ccccc|c}
\hline
Methods 
& Params (M) & FLOPs (T)  
& TotalSeg\_organs & TotalSeg\_vertebrae 
& TotalSeg\_cardiac & TotalSeg\_muscles
& TotalSeg\_ribs & TotalSeg\_all \\ \hline
nnU-Net~\cite{isensee2021nnu}         
& 31.28 & 0.54 
& 87.45 & 86.97
& 88.70 & 85.05
& 86.11 & 86.76
\\
nnU-Net*~\cite{isensee2021nnu} 
& 60.18 & 0.55
& 87.48 & 86.87
& 88.43 & 86.42
& 85.98 & 86.94
\\
nnFormer~\cite{zhou2021nnformer}
& 153.97 & 2.04
& 79.26 & 73.87
& 75.96 & 74.97
& 74.03 & 75.37
\\   
UNETR-S ~\cite{hatamizadeh2022unetr}
& 93.01 & 0.16
& 72.59 & 71.94
& 72.67 & 66.20
& 73.06 & 71.27
\\
UNETR-B ~\cite{hatamizadeh2022unetr}
& 102.02 & 0.59
& 77.11 & 74.21
& 77.57 & 73.29
& 74.06 & 75.05
\\
SwinUNETR-S ~\cite{hatamizadeh2022swin}
& 15.71 & 0.19 
& 76.29 & 77.04
& 76.05 & 71.33
& 74.21 & 74.94
\\
SwinUNETR-B ~\cite{hatamizadeh2022swin}
& 62.19 & 0.76 
& 84.18 & 83.27
& 83.45 & 81.07
& 81.71 & 82.64
\\
SwinUNETR-L$^\dagger$ ~\cite{hatamizadeh2022swin}
& 248.1 & 2.99 
& 85.39 & 87.27
& 85.25 & 82.97
& 83.63 & 84.91
\\ \hline
STU-Net-S 
& 14.60 & 0.13 
& 84.72 & 83.05
& 84.80 & 82.92
& 83.67 & 83.74
\\
STU-Net-B
& 58.26 & 0.51
& 87.67 & 86.46
& 88.64 & 86.46
& 86.82 & 87.12
\\
STU-Net-L
& 440.30 & 3.81
& 88.92 & 88.71
& 89.66 & 87.61
& 88.79 & 88.71
\\
STU-Net-H
& 1457.33 & 12.60 
& \textbf{89.82} & \textbf{90.43}
& \textbf{90.89} & \textbf{88.83}
& \textbf{90.29} & \textbf{90.06}
\\ \hline
\end{tabular}}
}
\end{table*}

\subsection{Quantitative Results on TotalSegmentator}
To verify the effectiveness of our STU-Net with different scales, we compare our models with other state-of-the-art methods on the validation set of TotalSegmentator. For a fair comparison, we train all models in nnU-Net framework for 1000 epochs. To improve the performance of other methods, we use the optimizers, learning rates, and learning rate decay strategies reported in their paper. Following the approach in the original SwinUNETR \cite{hatamizadeh2022swin} paper, where SwinUNETR-B has twice the feature size of SwinUNETR-S, we further scale the feature size of SwinUNETR-B by doubling it to 96, resulting in SwinUNETR-L. This allows us to observe the performance of SwinUNETR when the model is scaled up to larger size.

As shown in Figure \ref{fig:bubble} and Table  \ref{table:TotalSeg_Results}, our STU-Net-B model surpasses both the best CNN-based model, nnU-Net, and the best transformer-based model, SwinUNETR-B, by 0.36\% and 4.48\%  in terms of mean DSC on all classes, respectively. Further scaling our base model to large and huge sizes leads to 1.59\% and 2.94\% improvement in mean DSC score, respectively. We also observe that scaling up the SwinUNETR model leads to significant performance improvement, but still performs worse than our STU-Net-B. Our STU-Net-H achieves the highest mean DSC across all classes and within five sub-class groups in TotalSegmentator dataset. The results show the effectiveness of our architectural refinements on nnU-Net and the scaling strategy.

\subsection{Transferability of Trained Models}
We evaluate the transferability of our trained models by 1) conducting direct inference on the downstream CT datasets, which contain a subset of the 104 classes found in the TotalSegmentator dataset; and 2) fine-tuning the trained models on three downstream datasets, which include classes (e.g., lesions) not present in TotalSegmentator dataset and modalities (e.g., MR, PET) other than CT.

\subsubsection{Direct Inference Results}
We use the pre-trained STU-Net to conduct direct inference on 14 downstream datasets that include annotation categories within the TotalSegmentator dataset. These 14 datasets contain 2494 cases in total which is a strong benchmark to evaluate models' transferability pre-trained on large-scale datasets. 
 
The TotalSegmentator dataset provides detailed annotations for normal organs, including separate labels for left and right organs (without lesion annotations). However, downstream datasets may contain annotations for lesions within organ regions or combined annotations for left and right organs. These differences in annotation protocols can introduce extra labels that never appear in TotalSegmentator, leading to label inconsistency. To deal with this issue, we merge lesion annotations with their corresponding organ labels and, when necessary, combine left and right organ annotations to create modified labels for evaluation purposes. 

Table \ref{table:Inference_Results} shows that with TotalSegmentator for pre-training, models of larger scales usually have higher mean DSC scores across all these 14 datasets. This conclusion generally applies to each specific dataset. The better performance supports that our elaborately designed large-scale STU-Net pre-trained on the large-scale dataset can have better transfer capacities. 

\begin{table}[htbp]
\caption {Evaluation on the transferability of models trained on TotalSegmentator. Mean DSC ($\%$) of direct inference on various downstream datasets are reported. AMOS-CT denotes the results on the CT modality only.
}\label{table:Inference_Results}
\centering
\resizebox{\columnwidth}{!}{
{\begin{tabular}{c|c|c|c|c}
\hline
 Datasets &  nnU-Net\cite{isensee2021nnu,wasserthal2022totalsegmentator} & STU-Net-B &  STU-Net-L  &  STU-Net-H  \\ \hline
MSD Liver \cite{antonelli2022medical}
& 95.29 & 95.80 & 95.87 & 95.88 \\
MSD Pancreas \cite{antonelli2022medical}
& 76.52 & 75.96 & 78.41 & 78.95 \\
MSD Spleen \cite{antonelli2022medical}
& 93.97 & 95.46 & 95.38 & 95.52 \\
BTCV \cite{landman2015miccai}
& 80.14 & 80.96 & 83.05 & 83.83 \\
BTCV-Cervix \cite{landman2015miccai}
& 54.22 & 89.43 & 89.95 & 89.79 \\
AbdomenCT-1K \cite{ma2021abdomenct}
& 90.29 & 91.61 & 92.15 & 92.27 \\
KiTS2021 \cite{heller2020state}
& 86.00 & 83.45 & 84.38 & 85.44 \\
FLARE22\cite{MA2022102616} 
& 85.39 & 88.27 & 89.61 & 89.87  \\
CT-ORG \cite{rister2020ct}
& 69.01 & 72.09 & 71.90 & 73.14 \\
AMOS-CT \cite{ji2022amos}
& 79.26 & 80.99 & 82.87 & 83.11 \\
KiPA2022 \cite{he2021meta}
& 30.72 & 53.77 & 67.04 & 78.44 \\
Verse2020 \cite{liebl2021computed}
& 67.82 & 62.01 & 65.35 & 66.65 \\
WORD \cite{luo2022word}
& 78.73 & 76.07 & 78.08 & 77.42 \\
SegThor \cite{lambert2020segthor}
& 81.79 & 84.02 & 85.59 & 85.91 \\ \hline
Mean DSC (\%) 
& 76.37 & 80.71 & 82.83 & 84.02 \\
\hline
\end{tabular}}
}
\end{table}

\begin{table*}[htbp]
\caption {Fine-tuning results on 3 downstream datasets. Mean DSC ($\% \uparrow$) is evaluated. AMOS-CT (or -MR) denotes the results on the CT (or MR) modality only, otherwise, the results are on the mixed modalities. The same applies to AutoPET. The suffix 'ft' means fine-tuning. }\label{table:Finetune_Results}
\centering
\resizebox{\linewidth}{!}{
{\begin{tabular}{c|ccccccc|c}
\hline
Methods & FLARE22\cite{MA2022102616} & AMOS\cite{ji2022amos} & AMOS-CT\cite{ji2022amos} & AMOS-MR\cite{ji2022amos} & AutoPET\cite{gatidis2022whole} & AutoPET-CT\cite{gatidis2022whole} & AutoPET-PET\cite{gatidis2022whole} & Mean DSC (\%)  \\ \hline
nnU-Net\cite{isensee2021nnu}
& 85.86 & 89.48 & 89.98 & 87.39 & 73.03 & 41.84 & 71.81 & 77.06 \\
STU-Net-B        
& 86.56 & 89.70 & 89.84 & 86.92 & 73.96 & 41.09 & 72.25 & 77.19 \\
STU-Net-L       
& 87.41 & 90.29 & 90.57 & 87.04 & 74.94 & 40.97 & 73.37 & 77.80 \\
STU-Net-H        
& 87.29 & 90.13 & 90.39 & 87.18 & 72.19 & 43.72 & 72.59 & 77.64 \\ \hline
STU-Net-B-ft     
& 87.11 & 89.73 & 90.25 & 87.49 & 74.68 & 50.03 & 75.37 & 79.23 \\
STU-Net-L-ft    
& 88.25 & \textbf{90.50} & 90.79 & 87.69 & 75.19 & 51.88 & 77.04 & 80.19 \\
STU-Net-H-ft     
& \textbf{88.81} & 90.49 & \textbf{91.16} & \textbf{87.98} & \textbf{76.29} & \textbf{52.80} & \textbf{77.32} & \textbf{80.69}\\
\hline
\end{tabular}}
}
\end{table*}

\subsubsection{Fine-tuning Results}
We fine-tune the pre-trained STU-Net with different scales on three downstream datasets that contain novel anatomical structures, modalities and domains: AutoPET22 \cite{gatidis2022whole} (with lesions and PET modality), AMOS22 \cite{ji2022amos} (with MR modality), and FLARE22 \cite{MA2022102616} (with multiple domains). Owing to our architectural refinement for nnU-Net, all the model weights, except these of the segmentation head, can be easily transferred to these downstream tasks. 
Note that these downstream datasets, such as AutoPET, may contain multi-modal inputs which can have more channels than the input data of the pre-training dataset (e.g., a single channel for TotoalSegementator). More input channels require the first convolutional layer of the model to have more channels correspondingly, further resulting in a shape mismatch of convolutional weights between the pre-training and downstream fine-tuning tasks. In this case, we replicate the channels of the first layer's weights of the pre-trained model from a single channel to multiple channels to adapt to downstream tasks.
On AutoPET and AMOS datasets where multiple modalities exist, we fine-tune and evaluate our models on every single modality separately (denoted with a suffix of modality name in Table \ref{table:Finetune_Results}) and on the mixed modalities.

\begin{figure*}[htbp]
  \centering
   \includegraphics[width=\linewidth]{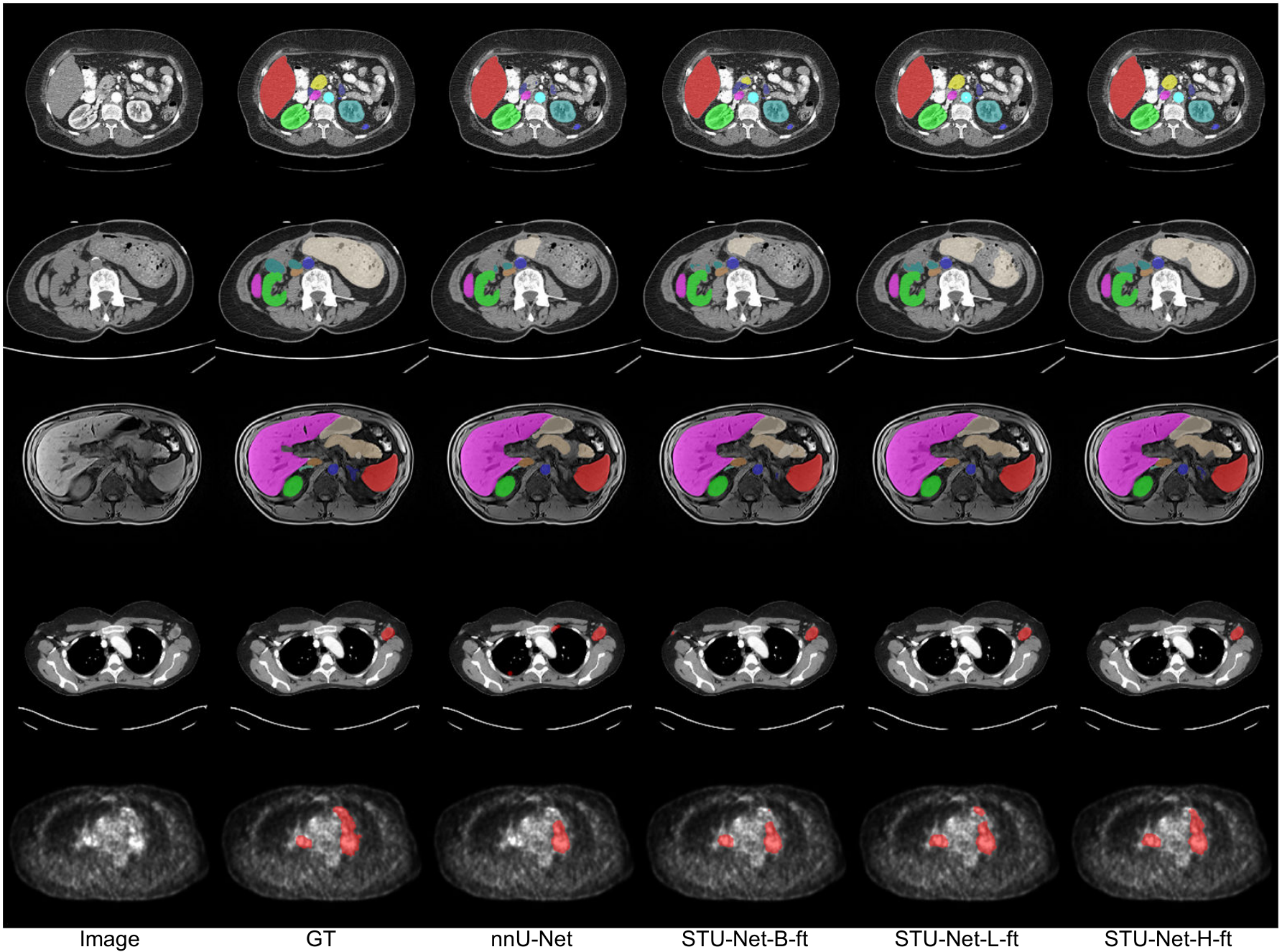}
   \caption{Qualitative visualization of our STU-Net with different scales and nnU-Net on various medical imaging datasets. The representative cases from distinct datasets are displayed in each row, including Row 1 - FLARE22 dataset, Row 2 - AMOS dataset with CT images, Row 3 - AMOS dataset with MR images, Row 4 - AutoPET dataset with CT images, and Row 5 - AutoPET dataset with PET images. The seven columns from left to right correspond to the original image, the ground truth (gt), the nnU-Net results, and our STU-Net-B-ft, STU-Net-L-ft, and STU-Net-H-ft results.}
   \label{fig:qualification}
\end{figure*}

As shown in Table \ref{table:Finetune_Results}, fine-tuning our STU-Net models, which are pre-trained on TotalSegmentator, leads to better segmentation performance than models trained from scratch on downstream datasets. Particularly, our huge model (STU-Net-H-ft) outperforms all the other models, achieving the highest mean DSC of 80.69\% on these downstream datasets. This observation underscores the importance of both pre-training and large model size in enhancing segmentation performance. A qualitative comparison of the segmentation results from our models and nnU-Net is provided in Figure \ref{fig:qualification}. This visual representation further highlights the advantages of our STU-Net models when fine-tuned on downstream datasets, and illustrates the superiority of our approach in various medical imaging scenarios.
Notably, the improvement over the other competitors is most significant on the AutoPET dataset, probably because pre-training on TotalSegmentator provides models with the information on the normal anatomy of the whole body. Such information is complementary to (and can be the prior information as) the lesion information which is the only annotation information in AutoPET. Thus, fine-tuning the pre-trained model can effectively enhance the model's ability to segment lesions on AutoPET. 

Interestingly, when trained from scratch, our huge model is slightly worse than the large one, probably because the limited training samples on these datasets cannot further boost the larger model. With sufficient training samples from TotalSegmentator for pre-training, our huge model can significantly benefit from fine-tuning and exceed the large one, usually by a clear margin. 

Notably, pre-trained models fine-tuned on non-CT modalities also demonstrate significant performance improvements, such as AMOS-MR and AutoPET-PET datasets, despite TotalSegmentator's focus on CT scans. This suggests that pre-training facilitates learning fundamental features and structures that generalize across modalities, beyond modality-specific characteristics.

\begin{table}[tbp]
\caption{Segmentation performance of models with different architectural variants on TotalSegmentator validation dataset. Mean DSC ($\% \uparrow$) is evaluated. *: We change the maximum feature number in standard nnU-Net from 320 to 512 to match the parameters of our STU-Net-B. Abbreviations: DS (Downsampling), US (Upsampling). }\label{table:ablation_arch}
\centering
\resizebox{\columnwidth}{!}{
\begin{tabular}{l|cc|c}
\hline
Architecture  & Params (M) & FLOPs (T) & Mean DSC (\%) \\
\hline
nnU-Net  & 31.28 & 0.54 & 86.76 \\
nnU-Net*  & 60.18 & 0.55 & 86.94 \\
\textbf{STU-Net-B}  & \textbf{58.26} & \textbf{0.51} & \textbf{87.12} \\
STU-Net-B (Replace w/ Conv DS) & 66.02 & 0.54 & 86.41 \\
STU-Net-B (Replace w/ Transpose Conv US) & 61.32  & 0.56 & 86.96 \\
STU-Net-B (Replace w/ Trilinear Interpolation US) & 58.26 & 0.51 & 86.67 \\
\hline
\end{tabular}
}
\end{table}

\subsection{Ablation Study}
\subsubsection{Efficacy of Architectural Refinements}
Our proposed STU-Net-B model is an improved version of the default nnU-Net architecture, incorporating several refinements. Table \ref{table:ablation_arch} compares the segmentation performance of different STU-Net-B architectural variants on the validation set of TotalSegmentator. Our STU-Net-B model utilizes the nearest interpolation for upsampling and incorporates downsampling in the first residual block. To evaluate its performance, we conducted a comparative analysis with alternative downsampling methods employing separate convolutions, and alternative upsampling methods utilizing transpose convolutions or trilinear interpolation.

We first increase the maximum feature number of standard nnU-Net from 320 to 512 to match the parameters of our STU-Net-B, and denote it as nnU-Net*. The nnU-Net* works better than the standard nnU-Net, but performs slightly worse than our STU-Net-B. This comparison demonstrates the effectiveness of the refinements in STU-Net-B. 

Then, we explore the downsampling design in STU-Net by introducing a variant that employs convolutional downsampling instead of integrating the downsampling process within the first residual block of each stage. This modification results in reduced performance and higher computational costs.
We further investigate the upsampling refinements by designing two variants of STU-Net-B. The first one utilizes transpose convolution to replace the default interpolation and convolution-based upsampling. This incurs a 0.16\% decrease in performance, and makes the weights not transferable for downstream fine-tuning. The second one uses trilinear (or cubic linear) interpolation to replace the nearest neighbor interpolation. This change decreases the performance and slows down the running speed. Overall, the default upsampling design in STU-Net achieves better performance, faster running speed, and better transfer capacities.

Thus, our proposed refinements not only enhance the effectiveness and efficiency of nnU-Net, but also endow it with the crucial properties of weight transferability and scalability. These properties are essential for further scaling up the model and facilitating transfer learning.

\subsubsection{Scaling Strategy}
We apply three different scaling strategies to nnU-Net and our STU-Net-base, namely scaling using different depth coefficient $d\in [1.0, 2.0, 3.0, 4.0]$, width coefficient $w\in [1.0, 2.0, 3.0, 4.0]$, and simultaneously with depth and width coefficients of $[1.0, 2.0, 3.0]$. The coefficient $d$ (or $w$) means that the depth (width) of each stage is scaled to $d$ (or $w$) times. 
Table \ref{table:ablation_scale} shows the results of different scaling strategies on the TotalSegmentator dataset. Firstly, wider nnU-Net* and STU-Net consistently obtain better performance while deeper ones are not. Therefore, compared to depth scaling, width scaling is more effective in improving the model performance on the large-scale dataset. But it brings a significant increase in computational consumption. Secondly, compared to the performance drop by deeper nnU-Net, our STU-Net can better benefit from scaling depth to a certain extent, e.g., 87.12$\rightarrow$87.72$\rightarrow$87.99$\rightarrow$87.58 vs. nnU-Net's 86.94$\rightarrow$85.65$\rightarrow$83.70$\rightarrow$81.45. We owe the better performance of STU-Net to its residual design. Thirdly, compound scaling, i.e., increasing depth and width simultaneously, is more effective and efficient in improving the performance of our STU-Net than the other two scaling strategies. Lastly, even with the same scaling strategy, and similar parameters and FLOPs, our STU-Net consistently outperforms nnU-Net* on all the settings, which again verifies the effectiveness of our refinement.

\begin{table}[tbp]
\caption {Segmentation performance of models with different scaling dimensions on TotalSegmentator dataset. Mean DSC ($\% \uparrow$) is evaluated. *: We change the maximum feature number in standard nnU-Net from 320 to 512 to match the parameters of our STU-Net-B.}\label{table:ablation_scale}
\centering
\resizebox{\columnwidth}{!}{
{\begin{tabular}{l|cc|c}
\hline
 Methods  & Params (M)  &FLOPs (T)  &  Mean DSC (\%)  \\ \hline
nnU-Net* (d=1.0, w=1.0)                  & 60.18 & 0.55 & 86.94  \\
STU-Net (d=1.0, w=1.0)                   & 58.26 & 0.51 & 87.12  \\ \hline
scale nnU-Net* by width (w=2.0)          & 240.47 & 2.19 & 87.29 \\
scale STU-Net by width (w=2.0)           & 232.80 & 2.00 & 88.08  \\ 
scale nnU-Net* by width (w=3.0)          & 540.88 & 4.92 & 87.80  \\
scale STU-Net by width (w=3.0)           & 523.62 & 4.49 & 88.85  \\ 
scale nnU-Net* by width (w=4.0)          & 961.40 & 8.73 & 88.84  \\
scale STU-Net by width (w=4.0)           & 930.71 & 7.97 & 89.19  \\ \hline
scale nnU-Net* by depth (d=2.0)          & 112.06 & 1.01 & 85.65  \\
scale STU-Net by depth (d=2.0)           & 110.15 & 0.96 & 87.72  \\ 
scale nnU-Net* by depth (d=3.0)          & 163.94 & 1.46 & 83.70  \\
scale STU-Net by depth (d=3.0)           & 162.03 & 1.41 & 87.99  \\ 
scale nnU-Net* by depth (d=4.0)          & 215.83 & 1.91 & 81.45  \\
scale STU-Net by depth (d=4.0)           & 213.91 & 1.86 & 87.58  \\ \hline
nnU-Net* compound scale (d=2.0, w=2.0)   & 447.97 & 4.00 & 87.42  \\
STU-Net compound scale (d=2.0, w=2.0)    & 440.30 & 3.81 & 88.71  \\
nnU-Net* compound scale (d=3.0, w=3.0)   & 1474.59 & 13.03 & 87.50  \\
STU-Net compound scale (d=3.0, w=3.0)    & 1457.33 & 12.60 & 90.06  \\
\hline
\end{tabular}}
}
\end{table}

\subsubsection{Pre-training and Fine-tuning Strategy}
We perform empirical studies on the effectiveness of pre-training, and then on mirror augmentation used in the pre-training and fine-tuning stages. Finally, we study the training epochs for pre-training. 

Firstly, we study the effectiveness of large-scale pre-training. We use STU-Net-L pre-trained on the TotalSegmentator dataset and then fine-tuned on FLARE22 and AutoPET datasets to conduct such a study. Comparing the 1st (w/o pre-training) and 3rd (w/ pre-training) rows in Table~\ref{table:ablation_finetune}, we find that the model with pre-training outperforms the one without pre-training by 0.84\% and 3.67\%, respectively. Such better performance justifies the effectiveness of large-scale pre-training.

Secondly, we investigate the mirror augmentation used in pre-training and fine-tuning. With pre-training, the mirror is useful for improving the DSC on the downstream task as the mirror (3rd$\sim$5th rows) improves the DSC of the 6th row (no mirror) considerably. It is also worth noting that the mirror used in both pre-training and fine-tuning stages achieves the best result.

Finally, with the mirror incorporated in both pre-training and fine-tuning, we find that different pre-training epochs can also influence performance (see the last four rows). The best result is obtained at the 4k epochs which can ensure that the model is sufficiently trained for convergence.

\begin{table}[tbp]
\caption {Segmentation performance of different pre-training settings on STU-Net-L. Mean DSC ($\% \uparrow$) is evaluated. The first and second rows compare whether to use mirror data augmentation when training models from scratch (without pre-training). The third to sixth rows compare the performance of different combinations of mirror data augmentation to pre-training and fine-tuning. The last four rows investigate the effect of different pre-training epoch numbers on fine-tuning results.  }\label{table:ablation_finetune}
\centering
\resizebox{\columnwidth}{!}{
{\begin{tabular}{c|c|c|c|c}
\hline
 Pre-train epochs &  Pre-train mirror  &  Fine-tune mirror & FLARE22\cite{MA2022102616} &  AutoPET-PET\cite{gatidis2022whole}   \\ \hline
- & - & \Checkmark & 87.41 & 73.37 \\
- & - & \XSolidBrush & 85.79 & 73.59 \\ \hline
4k & \Checkmark & \Checkmark & \textbf{88.25} & \textbf{77.04} \\
4k & \Checkmark & \XSolidBrush & 87.65 & 76.34 \\
4k & \XSolidBrush & \Checkmark & 86.16 & 75.88  \\
4k & \XSolidBrush & \XSolidBrush & 86.41 & 73.90   \\ \hline
1k & \Checkmark & \Checkmark & 87.06 & 74.89 \\
2k & \Checkmark & \Checkmark & 87.63 & 76.89 \\
4k & \Checkmark & \Checkmark & \textbf{88.25} & \textbf{77.04} \\
8k & \Checkmark & \Checkmark & 87.71 & 76.87 \\
\hline
\end{tabular}}
}
\end{table}

\begin{figure}[tbp]
  \centering
   \includegraphics[width=\linewidth]{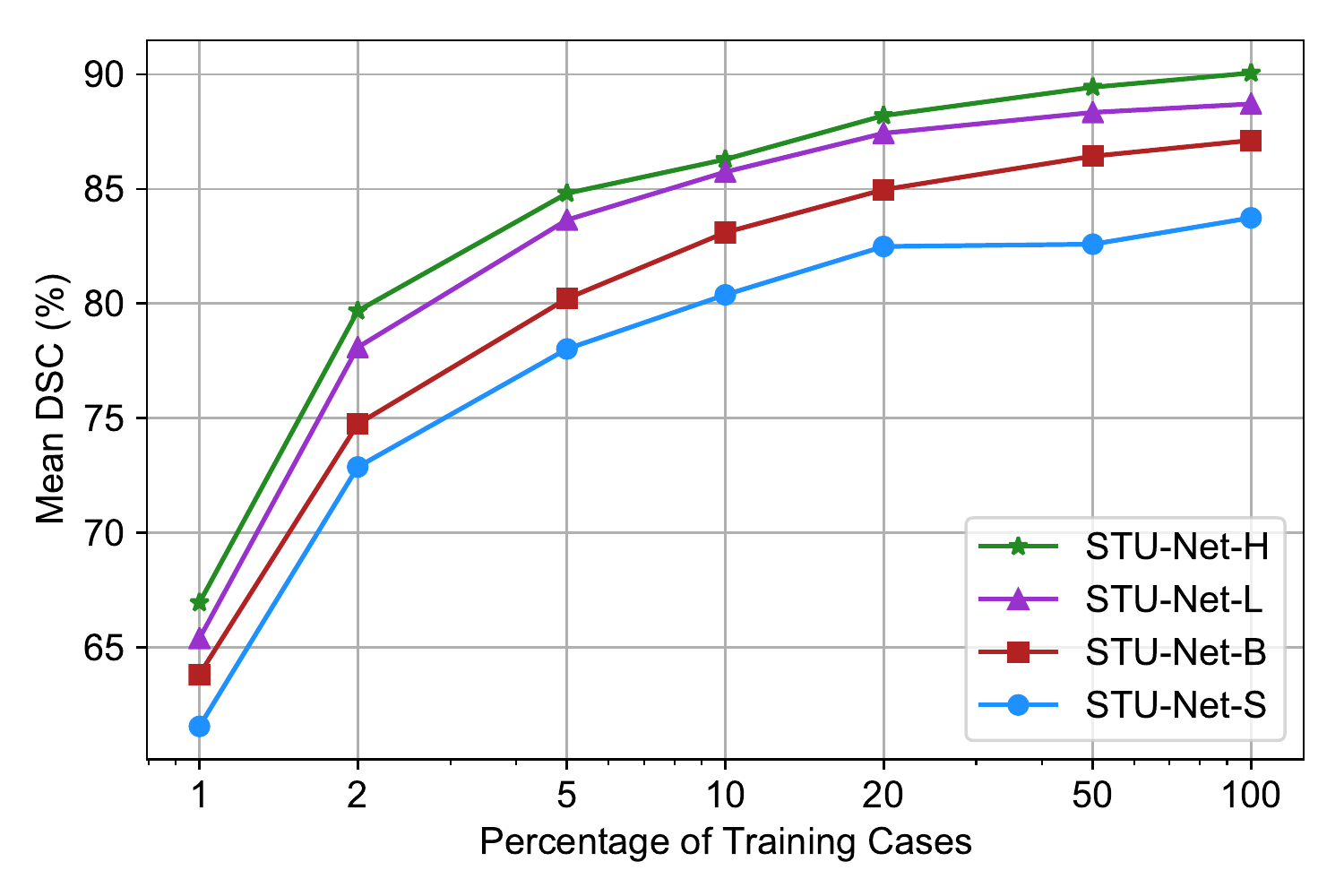}
   \caption{Comparison of mean DSC ($\% \uparrow$) performance for STU-Net models with different scales, trained on subsets of the TotalSegmentator training set with different proportions of training cases, and evaluated on the same TotalSegmentator validation set.}
   \label{fig:percent_case}
\end{figure}

\subsubsection{Influence of Dataset Sizes on Model Performance}
We investigate the impact of dataset size on the performance of our models when training them on the TotalSegmentator dataset. It is important to note that the training cases at different proportions were obtained through a stratified random selection process, ensuring that higher proportions of training cases also include the data from lower proportions.

As illustrated in Figure \ref{fig:percent_case}, increasing the model size leads to better segmentation performance on the TotalSegmentator subset, regardless of the number of training cases. For example, STU-Net-H outperforms STU-Net-S even when trained with only 5\% of the cases. Similarly, STU-Net-H surpasses STU-Net-B when trained with just 20\% of the cases. These results suggest that large-scale models are more data-efficient than smaller models for medical image segmentation. Moreover, the performance of different models is consistently improved with an increasing number of cases, and the trend has not yet saturated. These observations indicate that augmenting the number of training cases based on the TotalSegmentator dataset can yield further performance enhancements.

\begin{figure*}[tbp]
  \centering
   \includegraphics[width=\linewidth]{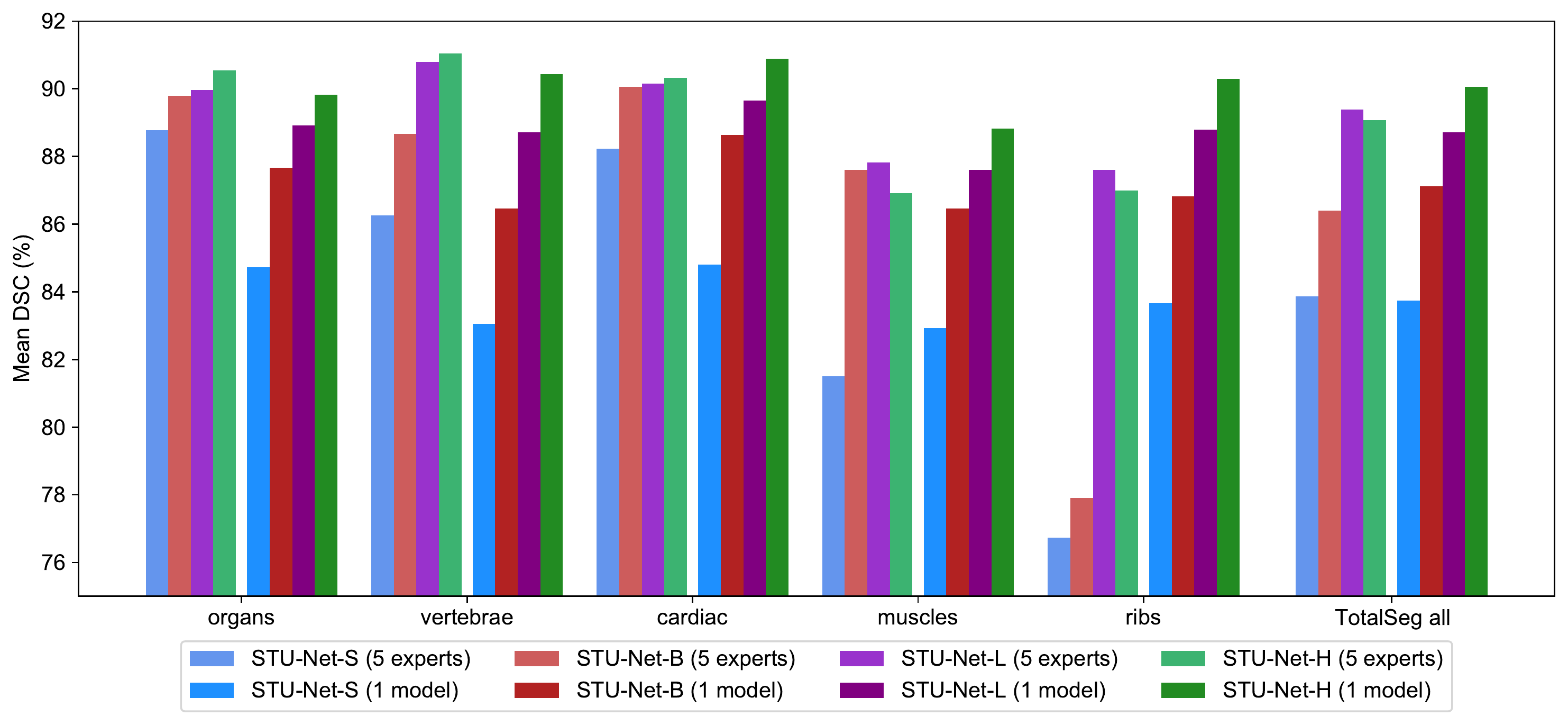}
   \caption{Comparison between five specialized expert STU-Net models and a single universal STU-Net model on the TotalSegmentator dataset. Each expert model targets one of the five subcategories (i.e., organs, vertebrae, cardiac, muscles, and ribs), while the universal model is trained on all 104 classes. The performance is measured using the mean DSC across various anatomical categories: the five subcategories and an overall performance metric for TotalSegmentator dataset. STU-Net architectures (S, B, L, and H) are depicted for both expert and universal models. Lighter colors represent expert models, and darker colors indicate universal models.}
   \label{fig:percent_expert}
\end{figure*}

\subsection{Universal Model Versus Expert Models}
We evaluate the performance of a universal STU-Net model trained on all 104 classes in the TotalSegmentator dataset, against five expert STU-Net models, each targeting one of the five subcategories (the same as in Table \ref{table:TotalSeg_Results}). Furthermore, we investigate the influence of model size on the performance of both expert and universal models.

Figure \ref{fig:percent_expert} illustrates that as model size increases, the performance of both expert and universal models generally improves. Expert models excel in the organs, vertebrae, and cardiac subcategories, while universal models perform better in the muscles and ribs subcategories. For the largest models (STU-Net-H), the universal model surpasses expert models, achieving the highest overall mean DSC score of 90.06\% on all the classes of TotalSeg dataset, compared to the expert models' highest mean DSC score of 89.07\%.

The results reveal that although expert models may outperform universal models in specific subcategories, universal models consistently deliver strong performance across different anatomical structures. The performance gap between expert and universal models varies across subcategories as the model size increases: the gap narrowing for organs and vertebrae, reversing for cardiac, and widening for muscles and ribs. The findings suggest that with increased model size, universal models are capable of concurrently segmenting numerous categories and exhibit promising performance advancements.

\section{Outlook}
Our work is an attempt toward Medical Artificial General Intelligence (MedAGI). We believe that the keys to MedAGI should be similar to that of AGI in the computer vision community -- the foundation model~\cite{bommasani2021opportunities,kirillov2023segment} and large-scale datasets. Large-scale datasets are emerging in the medical image processing field, while the foundation models, which can generalize to unseen tasks and data distributions, are less explored. The foundation models usually refer to the backbones which can be used to handle many tasks, like ResNet~\cite{he2016deep} or vision transformer~\cite{dosovitskiy2020vit,dehghani2023scaling} in computer vision. To this end, they are usually large-scale models with powerful representation abilities to extract discriminative features for different tasks. But in the medical image processing field, we lack such models. We thus explore how to design large-scale STU-Net models and take this problem as the initial step to the foundation models. Starting from STU-Net, researchers can build foundation models that can segment everything in medical images, like ~\cite{kirillov2023segment}, or further generalize these segmentation-based foundation models to many other tasks beyond segmentation, such as detection, classification, reconstruction and registration. Following this way, we will gradually approach the goal of foundation models: MedAGI, and use MedAGI to contribute to human health.

\section{Conclusions}
In this work, we introduce a series of scalable and transferable medical image segmentation models, named STU-Net, based on the nnU-Net framework. Our STU-Net models are designed to be scalable, with the largest model consisting of 1.4 billion parameters, making it the most substantial medical image segmentation model to date. By training our STU-Net models on the large-scale TotalSegmentator dataset, we demonstrate that scaling the model size yields significant performance improvements when transferring them to various downstream tasks. It thus underscores the potential of large-scale models in the medical image segmentation domain. 

Furthermore, our STU-Net-H model, trained on the TotalSegmentator dataset, exhibits robust transferability in both direct inference and fine-tuning scenarios across multiple downstream datasets. This observation emphasizes the practical value of leveraging large-scale pre-trained models for medical image segmentation tasks. 

In conclusion, the development of scalable and transferable STU-Net models has the potential to advance the state-of-the-art in medical image segmentation, opening new avenues for research and innovation within the medical image segmentation community.

%%%%%%%%% REFERENCES
{\small
\bibliographystyle{ieee_fullname}
\bibliography{egbib}
}

\newpage
\clearpage
\section*{Appendix}
\appendix
\begin{table*}[htbp]
\centering
\footnotesize
\begin{tabular}{lccl}
    \hline
    Datasets & \# Targets & \# Annotated Scans & Overlapped Annotations w/ Totalsegmentator \\
    \hline
01. MSD Liver \cite{antonelli2022medical}    & 1 & 131 & {Liver*} \\
02. MSD Pancreas \cite{antonelli2022medical} & 1 & 281 & {Pancreas*} \\
03. MSD Spleen \cite{antonelli2022medical}   & 1 & 41 & {Spleen} \\
04. BTCV \cite{landman2015miccai}            & 13 & 30 & {Spl, RKid, LKid, Gall, Eso, Liv, Sto, Aor, IVC, R\&SVeins, Pan, RAG, LAG} \\
05. BTCV-Cervix \cite{landman2015miccai}     & 1 & 30 & {Urinary Bladder} \\
06. AbdomenCT-1K \cite{ma2021abdomenct}      & 4 & 1000 & {Spleen, Kidney*, Liver, Pancreas} \\
07. KiTS2021 \cite{heller2020state}          & 1 & 300 & Kidney* \\
08. FLARE22\cite{MA2022102616}                 & 13 & 50 & {Liv, Eso, Sto, Duo, LKid, RKid, Spl, Pan, Aor, IVC, RAG, LAG, Gall} \\
09. CT-ORG \cite{rister2020ct}               & 4 & 140 & {Lung*, Liver, Kidney* and Bladder} \\
10. AMOS-CT \cite{ji2022amos}                & 15 & 200 & {Spl, RKid, LKid, Gall, Eso, Liv, Sto, Aor, IVC, Pan, RAG, LAG, Duo, Bla, Pro/UTE} \\
11. KiPA2022 \cite{he2021meta}                 & 1 & 70 & {Kidney*} \\
12. Verse2020 \cite{liebl2021computed}       & 24 & 61 & {Vertebrae C1-7, Vertebrae T1-12, Vertebrae L1-5} \\
13. WORD \cite{luo2022word}                  & 16 & 120 & {Spl, RKid, LKid, Gall, Eso, Liv, Sto, Pan, RAG, Duo, Col, Int, Rec, Bla, LFH, RFH} \\
14. SegThor \cite{lambert2020segthor}        & 3 & 40 & {Esophagus, Trachea, Aorta} \\
    \hline
\end{tabular}

\captionof{table}{
\textbf{Detailed information of 14 evaluation datasets for direct inference.} The table provides details on the number of overlapped targets, available annotated scans, and annotations that overlap with  TotalSegmentator.
*: Some conflict annotations underwent special processing to ensure consistency in targets, as described in Section \ref{sec:label_consist}.
}
\label{tab:public_dataset}
\end{table*}

\section{Consistency of overlapped annotations} \label{sec:label_consist}
In order to maintain consistency in the annotated targets, some  targets (e.g., Kidney \textit{v.s.} Left Kidney and Right Kidney) need to be processed due to variations in annotation protocols across datasets. In some cases, we combined the inference results of related target to ensure consistency. Specifically, the following results underwent special processing: \textbf{Kidney} was created by combining the results of Totalsegmentator targets \textit{kidney\_left} and \textit{kidney\_right}; \textbf{Lung} was created by combining five Totalsegmentator targets of lung lobe, i.e., \textit{lung\_lower\_lobe\_left}, \textit{lung\_lower\_lobe\_right}, \textit{lung\_middle\_lobe\_right}, \textit{lung\_upper\_lobe\_left}, and \textit{lung\_upper\_lobe\_right}. 

Additionally, considering the hierarchical relationships, the ground-truth annotations of organs and their inner tumors were combined. For the MSD Liver dataset, \textbf{Liver} was created by combining the ground truth of targets \textit{liver} and \textit{tumor}. Similarly, for the MSD Pancreas dataset, \textbf{Pancreas} was created by combining the annotations of targets \textit{pancreas} and \textit{tumor}. For the KiTS2021 dataset, \textbf{Kidney} was created by combining the annotations of targets \textit{kidney}, \textit{tumor}, and \textit{cyst}.

\section{Dataset Details}
In our evaluation, we utilized a total of 15 publicly available datasets, with 14 datasets used for direct inference and 3 datasets for fine-tuning. For direct inference evaluation, we conducted inference on all annotated cases and computed metrics for all overlapped targets between TotalSegmentator and the respective evaluation dataset, as detailed in Table \ref{tab:public_dataset}. Details on data splitting for fine-tuning will be introduced in the following parts.

\textbf{MSD Liver} \cite{antonelli2022medical} contains 131 contrast-enhanced CT scans annotated with liver and tumor. 

\textbf{MSD Pancreas} \cite{antonelli2022medical} consists of 281 portal venous phase CT scans annotated with pancreas and tumor. 

\textbf{MSD Spleen} \cite{antonelli2022medical} includes 41 3D volumes of CT with the annotation of spleen. 

\textbf{BTCV} \cite{landman2015miccai} consists of 30 cases of CT with labels of 13 abdominal organs. These images are collected from metastatic liver cancer patients or post-operative ventral hernia patients.

\textbf{BTCV-Cervix} \cite{landman2015miccai} includes 30 cervis CT scans annotated with 4 organs.

\textbf{AbdomenCT-1K} \cite{ma2021abdomenct} has 1000 cases of abdominal CT scans from 12 medical centers. Each case has annotations of 4 organs.

\textbf{KiTS2021} \cite{heller2020state} contains 300 annotated CT scans with labels of kidney, tumor and cyst. The data is collected from patients who underwent partial or radical nephrectomy for suspected renal malignancy between 2010 and 2020.

\textbf{CT-ORG} \cite{rister2020ct} is composed of 140 CT images containing 6 organ classes, which are from 8 different medical centers.

\textbf{KiPA2022} \cite{he2021meta} is provided from the Kidney PArsing Challenge 2022. This dataset contains 70 CT images with annotations of 4 kinds of kidney-related structures.

\textbf{Verse2020} \cite{liebl2021computed} is a CT spine dataset consisting of 374 scans from 355 patients. Considering the computational burden, we adopted the official train set~(61 cases) for evaluation.

\textbf{WORD} \cite{luo2022word} collects 150 CT scans from 150 patients before the radiation therapy in a single center. Each volume has 16 organs with fine pixel-level annotations and scribble-based sparse annotations. We conducted evaluation on all 120 available annotated cases.

\textbf{SegThor} \cite{lambert2020segthor} is the abbreviation of Segmentation of Thoracic Organs at Risk in CT images. SegThor provides 40 cases of CT scans with annotation of heart, trachea, aorta and esophagus. 

\textbf{FLARE22} \cite{MA2022102616} provides 50 cases of labeled CT images and 2000 unlabeled CT images. The segmentation targets include 13 organs. We tested all 50 annotated cases for direct inference. For fine-tuning, we trained on all labeled cases and evaluated on 20 official validation cases.

\textbf{AMOS} \cite{ji2022amos} has 500 CT scans (AMOS-CT) and 100 MR scans~(AMOS-MR) with voxel-level annotations of 15 abdominal organs. For our evaluation of direct inference, we utilized all 200 annotated scans from the training set of AMOS-CT. For fine-tuning, we trained on the training set~(200 CT and 40 MR) and evaluated on the validation set~(100 CT and 20 MR) according to official data splitting.

\textbf{AutoPET} \cite{gatidis2022whole} provides 1014 studies of paired 3D CT and PET scans from 900 patients where tumor lesions is annotated on each paired CT/PET images. Following the state-of-the-art solution \cite{ye2022exploring}, we fine-tuned our model on 398 cases and tested on 103 cases that contain lesions.

\end{document}